# A Comprehensive Study on Occlusion Invariant Face Recognition under Face Mask Occlusions

Susith Hemathilaka[1] and Achala Aponso[2]

[1]Department of Computer Science and Software Engineering,
University of Westminster, London, United Kingdom
[2]Department of Computer Science, Informatics Institute of Technology,
Colombo, Sri Lanka

*ABSTRACT*

*The face mask is an essential sanitaryware in daily lives growing during the pandemic period and is a big threat to current face recognition systems. The masks destroy a lot of details in a large area of face, and it makes it difficult to recognize them even for humans. The evaluation report shows the difficulty well when recognizing masked faces. Rapid development and breakthrough of deep learning in the recent past have witnessed most promising results from face recognition algorithms. But they fail to perform far from satisfactory levels in the unconstrained environment during the challenges such as varying lighting conditions, low resolution, facial expressions, pose variation and occlusions. Facial occlusions are considered one of the most intractable problems. Especially when the occlusion occupies a large region of the face because it destroys lots of official features.*

## 1. INTRODUCTION

Rapid development and breakthrough of deep learning in the recent past have witnessed the most promising results from face recognition algorithms but In uncontrolled environments face recognition systems drastically decrease their performance, due to various challenges such as facial expressions, pose variations, face occlusions, varying lighting conditions and low-resolution image inputs and different scales of images, partial captures, etc.[1]. Sometimes there are more than one challenges that make an incapable face recognition system detect and identify persons. During the pandemic, Occlusion problem is well highlighted when incapability of face recognition systems to recognize faces due to growing incentive of wearing a face mask to control spread of coronavirus. Even our mobile device's face recognition is incapable to identify the owners. This problem is more critically highlighted in airports, border control systems and security optimized premises.

Occlusions often happen in natural environments and they are most challenging and problematic in many fields of computer vision and object detection because any object can be occluded in the unconstrained environment and they destroy all details of the subject [2]. Because of that face recognition under occlusions remains a major challenge because of the unpredictable nature of occlusions. When it comes to facial recognition it's a very critical challenge because the occluded area varies in position, size and shape in face image [3]. Facial occlusions are unavoidable in unconstrained environments and there can be millions of different occlusion scenario which can't countable Practically, collecting a large dataset with all possible occlusion scenario to train a deep neural network is not feasible[4]. There are several scenarios which occur Facial occlusions can be categorized into four aspects[5].

- o Facial accessories: occluded by eyeglasses, facemasks, hat and hair.
- o External occlusions: occluded by hands or other random objects.
- o Partially captured faces: partially captured due to Limited field of view.
- o Artificial occlusions: occluded with random white or black rectangles, random salt & pepper noise.

## 2. FACE DETECTION

Face detection is the first step to face recognition which start the face recognition pipeline. General face detection approaches can be used up to some extent but there are some limitations in advanced unconstrained conditions. Detecting occluded faces is challengeable in unconstrained environments when a large area is occluded because it increases intraclass similarity and intraclass variations. Many approaches are developed using adaptive technologies to address the problem successfully.

### 2.1. General Face Detection

The existing face detection models less accurate sometimes when faces occluded. To address this problem, they introduced special algorithms for occluded face detection. MTCNN, Sckit-Image and Haar cascades can provide good results in a laboratory setting or indoor environments when detecting occluded faces. General face detection algorithms give an optimized performance in unconstrained environments as well. There are three main categories in face detection from approaches. Rigid templates, deformable part models and DCNN. The popular viola-jones face detection model, Harr like feature and AdaBoost comes under the rigid template category which can be drop performance in real-time applications[6]. DPM based suited better in real-time application, but their computational complexity is too much. Promising DCNN Approaches can solve the various problems that occurred in an unconstrained environment. DCNN provide a solid solution for various A-PIE problems up to date.

### 2.2. Face Detection under Occlusions

Occluded object detection is challengeable in unconstrained environments when a large area is occluded because of intraclass similarity and intraclass variations. Many approaches are taken to occluded object detection. Convolutional correlational filters are one of them to address the problem successfully. Binary segmentation is useful in the same task.

The existing face detection models less accurate in sometimes when faces occluded. To address this problem, they introduced special algorithms for occluded face detection. General face detection algorithms give an optimized performance in unconstrained environments. There are three main categories in face detection from approaches. Rigid templates, deformable part models and ConvNets. The popular viola-jones face detection model, Harr like feature and AdaBoost comes under rigid template category which can be drop performance in real-time applications. DPM based suited better in real-time application, but their computational complexity is too much. Promising DCNN Approaches can solve the various problems that occurred in an unconstrained environment [1]. DCNN provide a solid solution for various A-PIE problems up to date.

Handling Occlusions in faces are difficult because of occlusion variation and unknown locations. Efforts taken to detect occluded faces can be clustered into three main categories. Locating visible facial segments, Discarding the features taken from occluded sub-regions, Use the occlusion information. Attribute aware CNN like categorized face features according to the property of face part like big lips, long eye likewise to create facial response map. Most probably those approaches achieve good accuracies around 99%.FAN, LLE-CNN, AdaBoost cascade classifier are the approaches usually trained as segment-based face detectors from discarding occluded face sub-regions [2]. They achieve good accuracy, and their simple architecture makes them speed because of low computational complexity.

In the DCNN approaches, they are trying to minimize the occlusion damage by extracting features in the near area of occlusion. Novel grid-loss, AOFD, faster-CNN and LSTM hierarchical Attention Mechanism used to give better performance than other approaches.

## 3. FACE RECOGNITION

Facial recognition has been studied for several years and it's a longitudinal & preliminary task in computer vision[7]. Among the biometrics, methodologies face recognition has a better edge on the accuracy, efficiency, usability, security, and privacy to recognize the identity non-intrusively. Furthermore, Facial recognition is widely preferred over potential candidates because of its contactless authentication behaviour which helps to keep good hygiene practices[4]. Face recognition is widely used in personal device log-on, passport checking, ATMs, border control, forensics, law enforcement, surveillance, and national security[8].

### 3.1. Face Recognition Pipeline

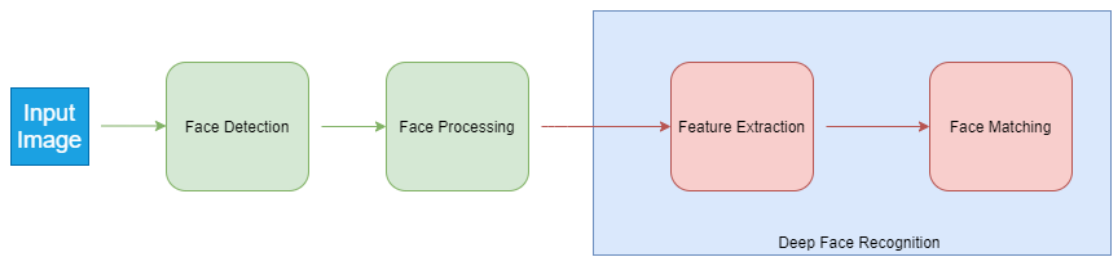

Figure 1. Face recognition pipeline

There are 4 components in the face recognition pipeline.

1. Face Detection- Detecting one or more faces in an image.
2. Face Processing- Cropping the face, scaling, and making alignments.
3. Feature Extraction- Extraction of important features from a face image.
4. Face Matching- Matching the similarity of extracted feature vector with the database   images.

### 3.2. Deep Face Recognition

Deep learning is an enabler of face recognition to achieve human-eye performance (Human: 97.53%) in recognizing faces in unconstrained environments. In 2014, DeepFace[9] achieved the SOTA accuracy of 97.35% in the LFW benchmark and rewarded as the first effort to reach the human-eye accuracy by training a 9-layer deep neural network on four million images. Afterwards, Face recognition accuracy surged to 99.80% due to advancements of deep learning-based approaches in just a few years. According to[5], the last five years obtain massive accuracy improvements when comparing the previous decade. The last four years recorded a 20x gain in accuracy achieved by advancements of deep learning approaches compared with early stages before 2014 which holds 4% of the failure rate.

## 4. OCCLUSION ROBUST FACE RECOGNITION

When Occlusion is in a small context and extracted features of the non-occluded area are more robust than occlusion is suitable for current face recognition systems. But the unpredictability of occlusion occurs problems in occlusion-robust approaches in practical environments. Occlusion robust approaches suggested new similarity measurement techniques and loss layers to deal with inter-class similarity. Those approaches try to keep the robustness at the same level when happening an occlusion but sometimes it's challenging. They try to leverage discriminative feature learning ability of face recognition systems. With the obtained success of deep learning approaches in face recognition systems handcrafted or engineering features are not satisfactory because those approaches are not adoptable into state-of-the-art systems. Learning-based features are the most suitable to address the problem in current deep learning face recognition systems. Facial descriptors are key players in handcrafted engineering-based approaches. These approaches are not realistic in practical situations; their poor alignment makes difficulties in feature extraction in a meaningful way. Local Binary Patterns and (SIFT) descriptor highly used in these applications. SIFT descriptors invariancy of illumination, pose, scale and rotation are very beneficial in practical environments.

The distance metric used in patch-based matching is important in those approaches. Elastic Bunch Graph Matching represents extracted facial features corresponding to the face representation of in a graph. Each node represents correspondence location features of the face and edges represent a relationship between different facial features in a related area. Learning-based features can be clustered into four. As appearance-based learning approaches, fisher face and eigenface are discriminately learn the subspace using in order PCA and LDA. Their limitation is proper alignment needed based on eye location. when eye location is occluded, they are not effective. Statistical learning approaches are introduced to account occlusion probability in different areas of the face. These approaches consider the knowledge of occluded facial parts [10]. introduced a scoring-based approach to select appropriate areas to validate to face recognition. They improved robustness of facial recognition systems considering the largest matching area of the occluded face. Sparse representation classifier leverages discriminative power than statistical learning. It combines linear combination and sparse errors to leverage discriminative power [11] study effort to overcome shortcomings of sparse representation approaches which perfectly works in laboratory conditions but dramatically fails in unconstrained environments. They introduce a structured occlusion coding approach to solve the limitation of the existing approach. They divide and recognize the occlusion separately to provide good classification to the probe image and reconstruction is learned by using mask strategy and a dictionary learning technique. Their work can be categorized into a robust face recognition approach than a recovery-based method. Their work has a limitation of collecting all occlusion scenarios that can happen in a practical environment and are not effective. When considering Discriminative power deep learning-based approaches are pioneered with comparing other learning-based approaches. The necessity of a lot of training data is a limitation of these approaches. Considering Deep learning approaches their architecture and loss functions to play an important role to leverage the robustness of the algorithms. Feeding a large dataset of collected occlusion can help relieve the occlusion problem but it's not practical and cost-effective. Some prior efforts have taken on the augmented face with occlusion used to train algorithms and they are not sufficient to improve the robustness into the satisfactory level. Solving data deficiency problems using data augmentation is not a practical solution and augmenting all the possible occlusion scenarios are inevitable.[12] the study found the sensitivity of the occlusion area to the facial recognition system. They conclude occlusion of the middle of the face is more difficult to handle. From this conclusion, they try to increase the discriminative power of the outer face to get better robustness. Researchers introduce some dictionaries that occlusion can happen in which area of the face and their effect according to occlusion location and it helps to identify occlusion patterns.

## 5. OCCLUSION RECOVERY FACE RECOGNITION

Most of the approaches attempt to solve the occlusion problem in the low level of face recognition pipeline which is known as feature space. Occlusion recovery approaches explicitly try to solve it at a higher level also known as image space. Occlusion recoveries have been done before to add data to face recognition algorithms. Then it becomes a more complex task than other approaches. There are several efforts and various approaches to recover clean faces from an occluded face. Reconstruction of the occluded area is another approach more competitive and better than inpainting. Space representation classification is the long-studied approach in reconstruction approaches. It overcomes the flaws of linear reconstructions efforts which used principal component analysis, recursive error compensation and Markov random field networks. It's an extension of linear combination by adding sparse errors accounting into the approach. SRC uses an identity matrix as an occlusion dictionary able to tackle occlusion precisely. Those approaches are not capable of contributing to current deep neural network-based systems. Few studies found which use deep learning techniques to tackle face de-occlusion problems. They usually learn encoding data from clean data and transfer them to corrupted or noisy data. Learnt decoding parameters are clarified to obtain clean data [13]. introduced the LSTM network and autoencoders to address shortcomings. Two LSTM components are introduced. One encodes face patches and other autoencoders decode the reconstruction representation. Adversarial ConvNets are capable to enhance the discriminative features more. Image inpainting is an emerging area in computer vision and deep learning. Those approaches are adopted to face de-occlusion problems as well.

Most studies focus on realistic image inpainting than their accuracy. Because several in painted approaches can't contribute to face recognition efforts. approaches can be clustered into two groups which are blind and non-blind inpainting considering the awareness of the location of corrupted pixels of an image. Most deep learning approaches are usually blind inpainting because deep learning is not able to discriminately learn the corrupted pixels like handcrafted approaches. In non-blinding inpainting, it copied mostly used textures in the image and replaced them in the occluded areas. In filling the pixels, the confidence value of the pixel is considered. As an extension of studies hybrid approaches are introduced to leverage face reconstruction. Finally, all recovery-based models are not provided with any promising results and improvements in face recognition evaluations.

There are a lot of new loss layers introduced in previous research to improve the similarity measurements on the feature extraction phase. Grid loss, SoftMax, Arc Face are some of them used in the latest facial recognition systems. Loss layers can improve the discriminative power of feature extraction and it improves the face recognition accuracy.

## 6. OCCLUSION DISCARD FACE RECOGNITION

Occlusion aware approaches are highly aware of the occlusion. Some of the approaches are discarded the occluded part of the face which is known as partial face recognition approach. Other Approaches considering the occlusion and try to minimize the effect of occlusion to face verification. We can divide Occlusion aware face recognition into two sections such as partial face recognition and context-aware feature extraction. Capturing Partial faces is often incident in the unconstrained environment due to occlusions and limited field of view or non-frontal faces. In security-critical situations such as surveillance identifying mugshots and criminal's investigation used it most. Some of the Occlusion based face recognition systems follow a similar approach to recognized faces can categorize into partially occluded face recognition. In this approach only qualifies the non-occluded area of the face to be used to face verification. Partial face matching is not satisfactory every time especially when occlusion free face appears. Partial face recognition approaches are not suitable in unified real-time systems due to its less efficiency and less-robustness to security breaches. [14] is a partial recognition approach using sparse representation classification combining to a Fully connected convolutional network to propose a novel approach called dynamic feature mapping (DFM). It can address partial faces regardless of size. They understood the necessity of partial face recognition in situations like criminal investigations and built the system for partial face recognition. Their work did not need fixed size and aligned facial images like others. Computational efficiency is high. But partially feature extraction is less secure when a holistic face is available. Furthermore, Partial face recognition needs a lot of training data and computational cost to train them. In our work, it discards non-discriminative features like jawline and chin shapes.

Robust face recognition has shown some influence in occluded face recognition. To date, it's not evaluated the capability of recognizing large occlusions like face masks. These approaches have some limitations when occlusion is varied, it challenges the system's robustness and usability. These approaches are more suited than partial face recognition in a unified face recognition system.[15] is a work introducing pairwise differential Siamese network (PDSN) with the intention for addressing the underlying problem of deep CNN's incapability of recognition occluded faces due to large intra-class variations and higher inter-class similarity caused by occlusions. With the inspiration of the human visual system works they detect occluded areas and discard them from face verification. They attempt to address the shortcomings of current face recognition systems when varying the occluded facial area on the face. They detect occlusion space in the image features using mask learning strategy to avoid corrupted feature vectors and introduced feature discarding mask and removed corrupted features by identifying correlations in top convolutional layers of corrupted face images. This work has given state-of-the-art results in general Occlusions.

## 7. DATA AUGMENTATION TOOLS FOR OCCLUSION GENERATION

[16] introduce tools for generating masked faces to existing face data by adding different types of masks. Their motivation comes because most face recognition systems already have in-house databases of faces that are not operable when the probe face is wearing a face mask. They avoid the threat of invalidating already collected databases of face recognition systems and without taking new pictures and recreating existing databases. It is a computer vision-based script to mask the face. They use DILB for facial key point detection to apply the mask to perfectly fit the face. It provides 100 different mask variations, and it can be used to convert any face dataset to masked face dataset. It supports multiple images in the same image, and it can bulk masking an entire dataset. They introduced a small dataset (MRF2) due to the lack of masked faces in datasets to retrain existing models and experimentally retrain existing face net models and reported the accuracy of face net improved up to 35% when wearing a face mask. It contains 53 identities and 269 images. They addressed the problem of invalidating existing databases of existing systems and evaluate performance improvement when training with masked faces.
[17] introduce three datasets for masked face detection (MFDD), masked face recognition (RMFD) and simulated masked face dataset (SMFRD). They identified the urgency of making a

Machine Learning and Applications: An International Journal (MLAIJ) Vol.8, No.4, December 2021

masked face dataset and introduced a real-world masked face dataset containing around 24,771 masked faces. MFDD mainly scrawled on the internet by using their introductory tool called RMFRD which can crawl the frontal facial images. It's capable of to train a masked face detection purposes and it used to detect when a person is wearing a mask or not as it is beneficial to control the epidemic situation. They introduce a tool to simulate and apply face masks to expand their dataset and add more diversity. The RMFD contains 5000 face images of 525 identities when wearing masks. We can consider their dataset as a good contribution to the addressing data insufficiency problem in masked face recognition.

[18] proposed a novel GAN based method to recreate the occluded area. They detect the masked face and then try to complete the image of the removed masked region. The first face they use binary segmentation to detect masked faces and then remove the area and synthesize. They use two discriminators to synthesize the face area, one discriminator for learning face structure and another to focus on learning the missing regions. Technically it finds similar patterns of facial features from a database of images and pastes in the occluded part. Their proposed approach removes masked areas automatically using gradually learning two discriminators. To address data insufficiency synthetical masked face dataset using CELEB-A face dataset. It's not qualitative enough for secure authentication and can't be preferable in real-time application because removing masked regions is not efficient.

## 8. EVALUATION TECHNIQUES

In generally, Face recognition adopts several evaluation metrics. Furthermore, it can use biometric authentication matrices and some of the common evaluation matrices from multi-class classification considering the nature of the solution.

### 8.1. Biometric Evaluation

In biometric evaluation, the authentication system always assumes a value in authentication between 0 and 1 meaning 1 is a full match and 0 is a no match. The threshold values play a vital and kept in a higher value should improve the robustness and security of the system. The threshold values should not be affected to the genuine user of the system[19]. The rate of not authenticating genuine users called False rejection rate (FRR) and authenticating the imposters are called False acceptance rate (FAR). The above rates are simplify using false positive, false negative, true negative and true positives as follows,

- FAR represents the rate of imposters that accepting as genuine users.

$$FAR = FP/(FP + TN)$$

- FRR indicates the rate of genuine users that recognized as imposters.

$$FRR = FN/(FN + TP)$$

- Weighted Error Rate is the weighted average of FAR and FRR.

$$WER = (FAR + FRR)/2$$

## 8.2. Classification Performance Evaluations

As mentioned earlier we can use general metrics to evaluate our algorithm performance as a multi class classifier. The evaluation techniques which can be used for the project as follows, the confusion matrix is used for evaluating the performance of classifiers models. It is relatively simple to understand and provide a good visualization of the model's performance. Confusion matrix built on four basic terminologies. Model accuracy is a clear indicator of model performance. Higher accuracy is better. Precision is the proportion of positive cases which was correct, and Recall is the proportion of actual positives that are identified properly. F1 score is the indication of the harmonic mean of precision and recall. The above metrics are calculated as below,

$$Accuracy = \frac{TP + TN}{TP + FP + TN + FN}$$

$$Precision = \frac{TP}{TP + FP}$$

$$Recall = \frac{TP}{TP + FN}$$

$$F1\ score = \frac{2 * precision * recall}{precision + recall}$$

## 9. IMPACT OF FACE MASK ON FACE RECOGNITION

Recognizing masked faces is considered as the most difficult facial occlusion challenge because it occludes a large area usually covering around 60% of the frontal face which contains rich features including the nose and mouth. Face mask makes higher inter-class similarity and inter-class variations due to covering a large area of the face which tricks the facial verification process of face recognition systems [17].

The study [20], concludes their results on the effect of wearing a face mask which given strong points of performance degrade on face recognition systems, showing the necessity of development of mask capable face recognition solutions. They evaluate the performance of the two state-of-the-art academic algorithms ArcFace [21] and SphereFace [22] and one non-academic algorithm which is state-of-the-art in face recognition performance. Their experimental setup is used to evaluate performance in face recognition without masks, when wearing a mask, with additional illumination (room light) and without illumination. Data taken under the mentioned four scenarios are used to evaluate three face recognition algorithms. Their results explicitly illustrated the degradation in the verification performance when considering masked face probes. These results illustrated the limitations of current face recognition solutions in matching masked probe faces with unmasked gallery faces.

Ongoing Face Recognition Vendor Test (FRVT) [5] is a performance evaluation report based on 89 face verification algorithms when wearing face masks [5] They baselined the performance of all algorithms using the original unmasked images and then applied synthetic masks digitally by varying mask shape, colour and coverage on acquired datasets from authorized travel or immigration processes. False rejection performance is highly affected when wearing face masks according to evaluation matrices. If masks cover around 70% of the face area most algorithms give false rejection rates between 20% and 50%. Minimum failure rate given by the quite competitive algorithm is 5% when face mask occlusion is less than 70%. In cooperative access control the personal log-on user can be prompted to the second attempt but it's not effective since the failure happens in algorithm level. Furthermore, false accepting rates 1 in 100000 impostors which is very critical in security consequences for verification.

## 10. DISCUSSION

Future dataset and analysis issues are addressed in this section. In certain cases, a modern study problem necessitates the use of complex datasets. Datasets, on the other hand, represent fundamental issues that must be addressed in the real world. Future challenges in the sense of dataset and research are mentioned in discussion.

The datasets have three major problems: dataset size, diversity of occlusions, and default standards. The occluded facial recognition datasets are on a small scale. AR is one of the few that has actual occlusions, with photographs of only 126 people included. Sunglasses and scarf occlusions are often regarded when it comes to occlusion diversity. Occlusions in real life, on the other hand, are far more varied. Unconstrained occluded facial detection will become a tricky challenge in the future that must be solved.

## 11. CONCLUSIONS

In this paper presents a comprehensive analysis on occluded face recognition strategies under face mask occlusion within detail comparisons, and we systematically classify approaches into occlusion robust, occlusion discard and occlusion recovery. Recently published and groundbreaking papers were discussed, including novel deep learning approaches. In addition, we demonstrate how face mask occlusions affect face recognition. Reader's attention is brought into room of improvements and future enhancement. Finally, we address upcoming dataset and analysis problems (along with possible solutions) that will help to advance the sector of face recognition under face mask occlusions.

## **AUTHORS**

**Susith Hemathilaka**, Final year undergraduate at University of Westminster, United Kingdom

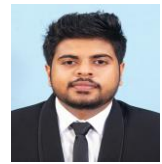

**Achala Aponso**, Senior lecturer at informatics institute of technology, Sri Lanka

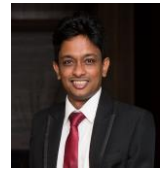